\documentclass[conference]{IEEEtran}
\IEEEoverridecommandlockouts
\usepackage{cite}
\usepackage{amsmath,amssymb,amsfonts}
\usepackage{algorithmic}
\usepackage{graphicx}
\usepackage{textcomp}
\usepackage{xcolor}
\def\BibTeX{{\rm B\kern-.05em{\sc i\kern-.025em b}\kern-.08em
    T\kern-.1667em\lower.7ex\hbox{E}\kern-.125emX}}
\begin{document}

\title{A Method for Waste Segregation using Convolutional Neural Networks}

\author{\IEEEauthorblockN{Jash Shah}
\IEEEauthorblockA{\textit{Department of Information Technology} \\
\textit{K J Somaiya College of Engineering}\\
Mumbai, India, \\
jash12@somaiya.edu}
\and
\IEEEauthorblockN{Sagar Kamat}
\IEEEauthorblockA{\textit{Department of Information Technology} \\
\textit{K J Somaiya College of Engineering}\\
Mumbai, India, \\
sagar.kamat@somaiya.edu}

}

\maketitle

\begin{abstract}
Segregation of garbage is a primary concern in many nations across the world. Even though we are in the modern era, many people still do not know how to distinguish between organic and recyclable waste. It is because of this that the world is facing a major crisis of waste disposal. In this paper, we try to use deep learning algorithms to help solve this problem of waste classification. The waste is classified into two categories like organic and recyclable. Our proposed model achieves an accuracy of 94.9\%. Although the other two models also show promising results, the Proposed Model stands out with the greatest accuracy. With the help of deep learning, one of the greatest obstacles to efficient waste management can finally be removed. 

\end{abstract}
\vspace{\baselineskip}

\renewcommand\IEEEkeywordsname{Keywords}

\section{Introduction}
The garbage and recycling businesses are being overwhelmed by the ever-increasing volume of global garbage. As a result, the demand for smart solutions for environmental monitoring and recycling process improvement is stronger than ever [1]. Human life and the environment are both affected by waste disposal, whether directly or indirectly. The negative consequences of waste materials can be mitigated with the use of a competent waste management system [2]. Currently, there are two types of waste classification and separation: manual waste classification and automated waste classification using multiple techniques. The first may be accomplished with human intelligence and power, while the second entails the automatic search for appropriate waste classification techniques [3].

Recycling is rapidly becoming an essential component of a sustainable society. However, the entire recycling process has a high hidden cost. This is caused by the recycling materials' selection, categorization, and processing. Even while many customers nowadays are able to perform their own garbage sorting, they may be puzzled about how to select the correct waste category when disposing of a wide range of items. Finding an automated approach to recycling is currently extremely valuable in today's industrial and information-based world since it offers both environmental and economic benefits [4].

Dumping organic wastes in landfills is a big concern, not because of the resources lost in the process, but because the organic waste undergoes anaerobic decomposition in the landfill, resulting in methane production. Methane has a greater greenhouse gas effect than carbon dioxide when discharged into the atmosphere. Organic waste, on the other hand, has its own set of issues, since it may be a source of greenhouse gases, methane, and pollution. If organic waste is not properly cleaned or controlled, it can infiltrate water sources and feed bacteria, resulting in the formation of fungus, which can be hazardous to society.

The majority of municipal trash is not segregated at the source of origin and ends up in sanitary landfills. That is why we require a more effective system for distinguishing recyclable materials, and this is where we can help. We can identify, remove, and sort those items on a moving conveyor using computer vision and artificial intelligence (AI). CNN (convolutional neural networks) characteristics are used by the majority of today's top-performing object detection networks. A more automated approach allows us to ship fewer recyclables to landfills. In this paper, we built three models, namely ResNet-34, VGG16 and a proposed deep neural network for image classification, segmentation, and detection and compared the results that we obtained. The categorization of recyclable and organic materials is a difficult topic that necessitates the use of advanced approaches. A worldwide strategy for industrial applications is required, in addition to dataset gathering. We offer a more accurate and optimum waste categorization approach in this study.

\section{Literature Survey}
In paper [5], the OpenCV computer vision library is used in this study to conduct data improvement, and picture preparation on the waste images gathered. A VGG16 convolutional neural network is created using TensorFlow as the model training backdrop, with the RELU activation function and the addition of a BN layer to enhance the rate of convergence of model and recognition accuracy rate. The correct rate of the scheme described in this paper is 75.6 per cent after testing on the test set. The scheme described in this paper can effectively classify domestic garbage into toxic garbage, kitchen waste, other garbage, and recyclable waste, meeting the needs of practical applications.

In paper [6], a comparison is made between a proposed model that is WasteNet and various other models like VGG, AlexNet, ResNet, DenseNet and SqueezeNet on four different metrics. The waste is categorised into six different categories. It was found that the proposed model performed better on all four metrics and had an accuracy of 97\%. Hence, it was then used to make a smart dustbin that classifies the waste that comes into the dustbin. 

The paper [7] proposes a neural network for waste classification into three categories: recyclable, non-recyclable and organic waste, with an accuracy of 81.22\%. The paper also compares other models like VGG16, Inception-Net, Dense-Net and Mobile-Net. Out of all these transfer learning models, Mobile-Net showed the highest accuracy of 92.65\%. 

Paper [8] suggests using a Convolutional Neural Network and a Support Vector Machine algorithm for better waste classification. They have used ResNet-50 as the transfer learning model as the dataset was small. They segregate waste into four categories: glass, paper, plastic and paper. By this approach, they have achieved an accuracy of 87\% on epoch number 12. 

Trashnet datasets were utilized to conduct this [9] study. The performance of the SVM classification using the SIFT feature is compared to the performance of the same technique using the SIFT-PCA combined feature. The results of the experiments indicate that categorization using SIFT feature extraction achieves a 62 per cent accuracy.

The authors of [10] look at a variety of approaches and give a thorough assessment. Support vector machines using HOG features, basic convolutional neural networks (CNN), and CNN with residual blocks were among the models we used. The authors concluded that basic CNN networks with or without residual blocks perform well, based on the results of the study.

The critical component of the system mentioned in [11] is a garbage container that will automatically sort waste using the Internet of Things and Machine Learning technologies. The bin is linked to the cloud, which aids in the waste collection by recording and uploading numerous data points for each bin. This study describes two versions of the system, the first of which achieves a 75 per cent accuracy in categorising garbage as wet or dry, and the second of which achieves a 90 per cent accuracy in sorting waste into six unique categories.

In paper [12], the performance of four CNN-based Waste Classifiers, namely ResNet-50, VGG16, MobileNet V2 and DenseNet-121. The waste was categorized into four categories like hazardous waste, general waste, recyclable waste and compostable waste. The highest accuracy of 94.8 per cent was obtained for ResNet-50. 

Using a deep learning technique, the paper [13] examines a unique approach for waste sorting for successful recycling and disposal. The YOLOv3 method was used to train a self-made dataset in the Darknet framework. The network has been taught to recognise six different categories of objects. The detection test was also done using YOLOv3-tiny for comparative assessment to evaluate the competency of the YOLOv3 algorithm.

The authors in paper [14] have classified the waste into four classes like cardboard, paper, metal and plastic. They designed a Convolutional Neural Network, which attained the highest testing accuracy of 76 per cent for 100 epochs for 50x50 size images.  

\section{Dataset}
Kaggle provided the dataset for the transfer learning model and the proposed CNN model. It is waste classification data, which comprises approximately 25,077 waste images split into two categories: organic and recyclable. Table I indicates the number of images in each class, whereas Fig. 1 depicts the same in a pie chart. Fig. 2 shows some sample images in the input dataset.

\begin{table}[htbp]
\caption{Number of waste images in the dataset.}
\begin{center}
\begin{tabular}{|l|l|l|}
\hline
              & Organic & Recyclable \\ \hline
Training data & 12565   & 9999       \\ \hline
Testing data  & 1401    & 1112       \\ \hline
Total         & 13966   & 11111      \\ \hline
\end{tabular}
\end{center}
\end{table}

\begin{figure}[htbp]
\centerline{\includegraphics{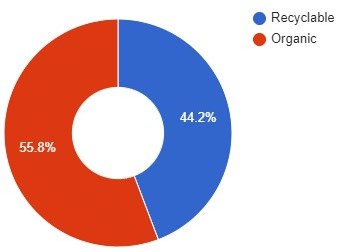}}
\caption{Distribution of waste images in dataset.}
\label{fig}
\end{figure}

\begin{figure}[htbp]
\includegraphics[width=8.5cm,height=14cm,keepaspectratio]{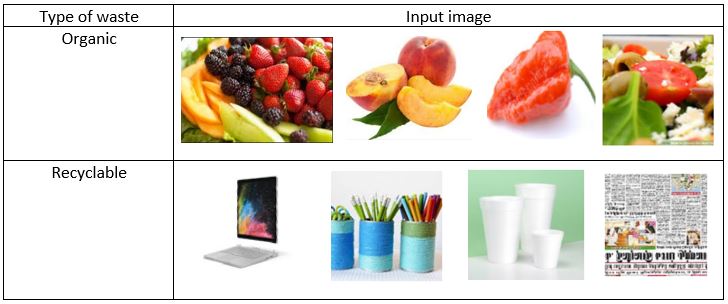}
\caption{Types of waste images in dataset.}
\label{fig}
\end{figure}

\section{Methodology}
\subsection{VGG16}
In 2014, Zisserman and Simonyan suggested the VGG16 network structure as one of the VGG NET networks [15]. It's a more advanced version of the AlexNet network. When recognising and categorising pictures, it can more correctly convey the features of the data collection. Large-scale data sets and complicated backdrop recognition tasks benefit from it. Thirteen convolutional layers, three fully connected layers, and five pool layers make up the network structure. The convolution kernel used in the 13 convolutional layers contained in VGG16 is a medium-sized 3 x 3 matrix with a moving step of 1 compared to other networks. The number of convolution kernels steadily grew from 64 in the first layer to 128 in the second, 256 in the third, and finally 512 in the final layer. The pooling layer's convolutional kernel is 2 x 2 in size, and the step size is 2. It performs better on the retrieved features than other networks with a convolution kernel size of 5 x 5. VGG model offers superior processing capabilities for training data sets with tiny amounts of data, and it is easier to deploy and has higher recognition accuracy than other deep convolutional neural networks [16]. 

The VGG16 network model, on the other hand, offers both advantages and downsides in terms of precise identification. The number of model parameters and the complexity of computations during training has risen as the network structure has been deepened, resulting in a long training period and low training efficiency. The following approaches are utilised to enhance the VGG16 model in this work in order to maintain the key features of the model extraction without lowering the accuracy of recognition while also improving the pace of model training and minimising the time it takes to train the model. 
The Relu function is used to activate the VGG 16 network in the VGG network, and the formula is:

\begin{equation}
f(x) = \begin{cases}
x & \text{ if } x\geq 0\\        
0 & \text{ if } x< 0
\end{cases}
\end{equation}

Where: x is the input of the Relu function; f(x) is the function's output. 
The loss function is used to measure the difference between the predicted and actual values throughout the model training process. In the VGG 16 network, use the CrossEntropyLoss loss function, which has the following formula:
\begin{equation}
    E(t,y)=-\sum_{j}^{}t_{j}\log y_{j}
\end{equation}

Among them: t and y represent the target label and output of the neural network, respectively, and represents the softmax loss function: 
\begin{equation}
y_{j} = e^{z_{j}}/\sum_{k}^{}e^{z_{k}}
\end{equation}

\subsection{ResNet - 34 (FastAI)}
ResNet-34 is a state-of-the-art image classification model with 34 layers of convolutional neural networks. This is a model that has been pre-trained on the ImageNet dataset, which has 100,000+ pictures divided into 200 categories [17]. It differs from traditional neural networks in the sense that it uses the residuals from each layer in the subsequent connected layers.

The infrastructure of the ResNet-34 network is the residual building component, and it makes up the majority of the network. The residual building block used a shortcut connection to skip the convolutional layers, effectively alleviating the problem of gradient disappearance or explosion characterized by increasing depth in neural networks and allowing us to build CNN structures more freely and improve the rate of recognition [18].

ResNet has one convolution and pooling step, the four layers of identical behaviour.
Each layer follows a similar pattern. They execute 3x3 convolution with a fixed feature map dimension (F) of [64, 128, 256, 512] and skip the input after every two convolutions. Furthermore, the width (W) and height (H) of the layer stay consistent throughout. [19]

\subsection{Proposed Model}
\subsubsection{Convolutional Layer}
The input image is first passed via a convolutional layer of the convolutional neural network. The convolutional layers use convolution to extract high-level characteristics from an input image. A convolution is a linear procedure in which the input is multiplied by a set of filters. The convolutional neural network, unlike previous algorithms, learns the filters from the training dataset. The earlier convolutional layers generally capture the lower-level features such as gradient orientation, edges, and so on. Increasing the number of convolutional layers enhances the ability to encode higher-level features. [20]

\subsubsection{Pooling Layer}
In the pooling layer, the output of the ReLU function is then dimension reduced. To extract the salient characteristics important in the feature maps, the pooling layer uses a max-pooling method. The maximum pooling layer's output is more rotational and translational invariant. Max pooling can also be used to filter out noisy and minor activations, lowering the number of computing resources required.

\begin{figure}[htbp]
\centerline{\includegraphics[width=8.5cm,height=13cm,keepaspectratio]{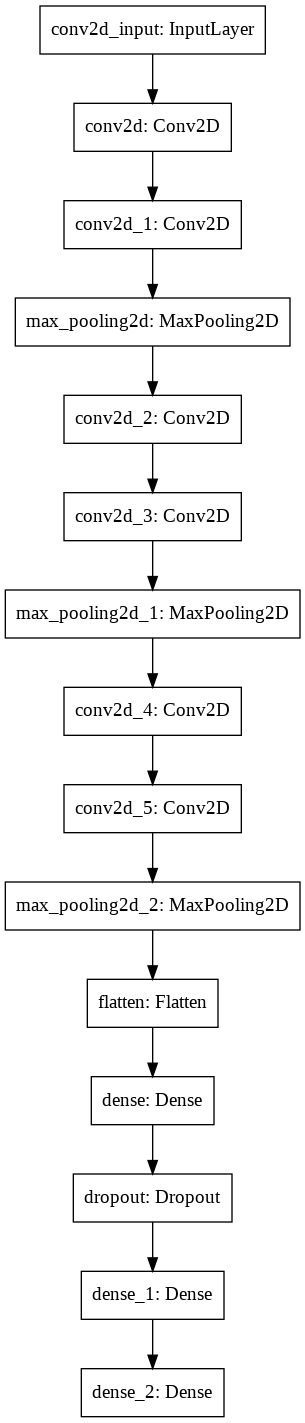}}
\caption{Architecture of proposed CNN model.}
\label{fig}
\end{figure}

\subsubsection{Fully Connected Layer}
The features are flattened and sent into the fully connected layer after going through the above stages. The fully connected layer receives the features from the previous layers and uses the backpropagation technique to learn the non-linear functions of the features. The last fully-connected layer uses a softmax function to calculate each class's probability. All of the classes' probabilities will add up to one.

\subsubsection{Activation Layer}
Before passing to the next layer, the output from the convolution layer is activated by a ReLU activation function. When compared to the commonly used sigmoid and tanh functions, the ReLU function has the advantage of just activating non-negative neurons, making it more computationally efficient.

In this work, the desired shape of input images is 224x224 with an RGB color scheme. The CNN used in this work contains 6 Conv2D layers, 3 MaxPool2D layers and three fully connected Dense layers. ReLU acts as the activation function in the fully connected layers. The output layer contains only a single neuron which will contain values as 0 or 1, where 0 stands for class ('Organic') and 1 for class ('Recycled').

\section{Results}
In ResNet-34 architecture with seven epochs, an accuracy of 91.8\% was achieved. For the VGG16 model, an accuracy of 93.37\% was obtained on training the model for five epochs and a batch size of 32.

\begin{table}[htbp]
\caption{Classification report on test data.}
\resizebox{0.48\textwidth}{!}{%
\begin{tabular}{|c|c|c|c|c|c|c|}
\hline
          & \multicolumn{2}{c|}{VGG16} & \multicolumn{2}{c|}{ResNet-34} & \multicolumn{2}{c|}{Proposed model} \\ \hline
          & Organic    & Recyclable    & Organic      & Recyclable      & Organic         & Recyclable        \\ \hline
Precision & 0.86       & 0.89          & 0.94         & 0.92            & 0.96            & 0.95              \\ \hline
Recall    & 0.92       & 0.81          & 0.93         & 0.91            & 0.94            & 0.96              \\ \hline
f1-score  & 0.89       & 0.85          & 0.94         & 0.93            & 0.95            & 0.94              \\ \hline
\end{tabular}%
}
\end{table}

\begin{figure}[htbp]
\centerline{\includegraphics[width=8.5cm,height=5cm]{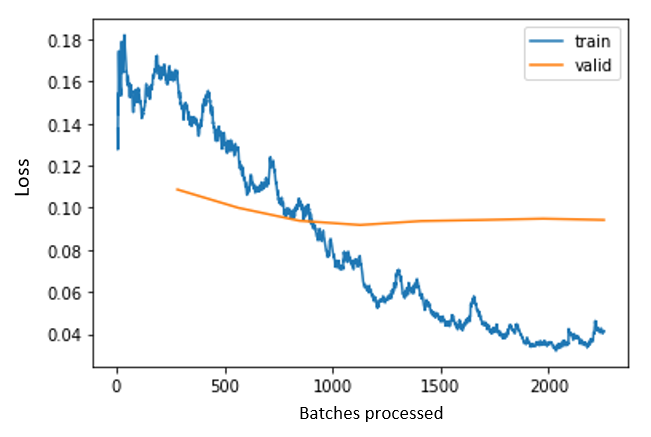}}
\caption{Evaluation of loss function of ResNet-34 model.}
\label{fig}
\end{figure}

\begin{figure}[htbp]
\centerline{\includegraphics[width=8.5cm,height=5cm]{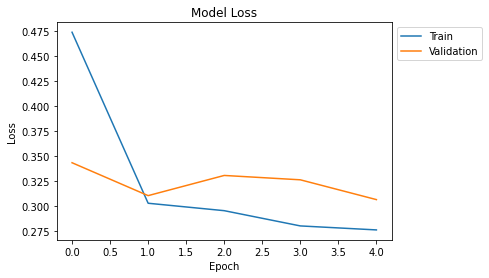}}
\caption{Evaluation of loss function of VGG16 model.}
\label{fig}
\end{figure}

\newpage
In the proposed model, the batch size was set to 64, and the number of epochs was kept at 10, but it was observed that the accuracy of the model was not improving after epoch number 7. Hence, an early stopping method was applied, and accuracy of 94.96\% was attained.

\begin{figure}[htbp]
\centerline{\includegraphics[width=8.5cm,height=12cm,keepaspectratio]{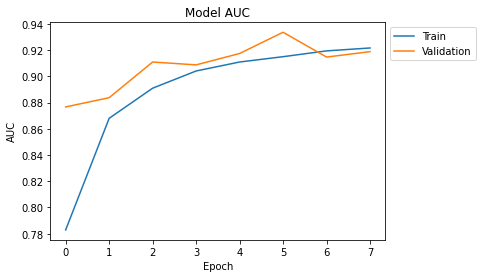}}
\caption{Evaluation of accuracy for proposed CNN model.}
\label{fig}
\end{figure}

\begin{figure}[htbp]
\centerline{\includegraphics[width=8.5cm,height=15cm,keepaspectratio]{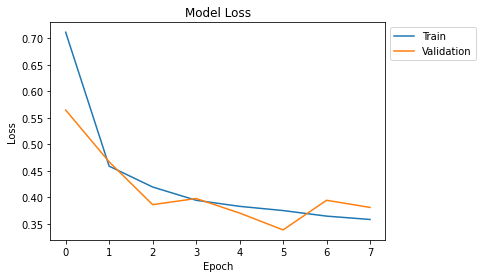}}
\caption{Evaluation of loss function for proposed CNN model.}
\label{fig}
\end{figure}

\begin{figure}[htbp]
\centerline{\includegraphics[width=8.5cm,height=15cm,keepaspectratio]{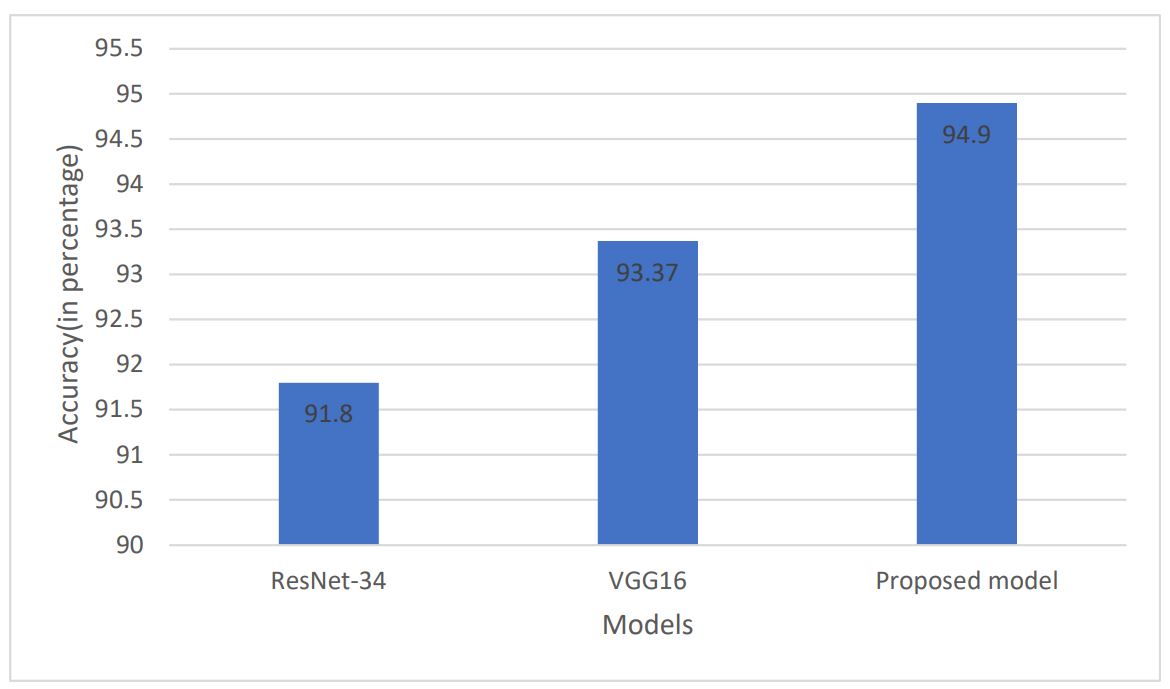}}
\caption{Bar chart representation of accuracies of various models.}
\label{fig}
\end{figure}

\newpage
\section*{Conclusion}
Segregation and management of waste have been a long-standing issue that has impacted a vast percentage of the ecosystem. It is easier to manage waste using today's technology if they are applied properly.

As we can see from the findings of this study, the problem of waste image categorization can be tackled with deep learning algorithms with a high degree of accuracy. It can thus be seen from the comparison of accuracies of the three deep learning models the proposed has the highest accuracy of 94.9\%. Though the accuracy of other models doesn't come at par with the proposed model, still these models can be considered for further improvement since the accuracy of those architectures is above 90\%.


\begin{thebibliography}{00}
\bibitem{b1} H. Wang, "Garbage Recognition and Classification System Based on Convolutional Neural Network VGG16," 2020 3rd International Conference on Advanced Electronic Materials, Computers and Software Engineering (AEMCSE), 2020, pp. 252-255, doi: 10.1109/AEMCSE50948.2020.00061.
\bibitem{b2} White, Gary et al. “WasteNet: Waste Classification at the Edge for Smart Bins.” ArXiv abs/2006.05873 (2020).
\bibitem{b3}Thanawala D., Sarin A., Verma P. (2020) An Approach to Waste Segregation and Management Using Convolutional Neural Networks. In: Singh M., Gupta P., Tyagi V., Flusser J., Ören T., Valentino G. (eds) Advances in Computing and Data Sciences. ICACDS 2020. Communications in Computer and Information Science, vol 1244. Springer, Singapore. https://doi.org/10.1007/978-981-15-6634-9\_14
\bibitem{b4} Olugboja, Adedeji \& Wang, Zenghui. (2019). Intelligent Waste Classification System Using Deep Learning Convolutional Neural Network. Procedia Manufacturing.35.607-612.10.1016/j.promfg.2019.05.086.
\bibitem{b5} A. P. Puspaningrum et al., "Waste Classification Using Support Vector Machine with SIFT-PCA Feature Extraction," 2020 4th International Conference on Informatics and Computational Sciences (ICICoS), 2020, pp. 1-6, doi: 10.1109/ICICoS51170.2020.9298982.
\bibitem{b6} S. Meng and W. -T. Chu, "A Study of Garbage Classification with Convolutional Neural Networks," 2020 Indo – Taiwan 2nd International Conference on Computing, Analytics and Networks (Indo-Taiwan ICAN), 2020, pp. 152-157, doi: 10.1109/Indo-TaiwanICAN48429.2020.9181311.
\bibitem{b7}Ziouzios, D.; Tsiktsiris, D.; Baras, N.; Dasygenis, M. A Distributed Architecture for Smart Recycling Using Machine Learning. Future Internet 2020, 12, 141. https://doi.org/10.3390/fi12090141
\bibitem{b8} Gao, M.; Qi, D.; Mu, H.; Chen, J. A Transfer Residual Neural Network Based on ResNet-34 for Detection of Wood Knot Defects. Forests 2021, 12, 212. https://doi.org/10.3390/f12020212
\bibitem{b9} K. He, X. Zhang, S. Ren and J. Sun, “Deep Residual Learning for Image Recognition,” in CVPR, 2016.
\bibitem{b10} K. Simonyan and A. Zisserman, “Very Deep Convolutional Networks for Large-Scale Image Recognition,” in arXiv:1409.1556 [cs], San Diego, CA, USA, pp. 1–14, May 2015, [Online]. Available: http://arxiv.org/abs/1409.1556 [Accessed: Oct. 24, 2021].
\bibitem{b11} S. Albawi, T. A. Mohammed and S. Al-Zawi, "Understanding of a convolutional neural network," 2017 International Conference on Engineering and Technology (ICET), 2017, pp. 1-6, doi: 10.1109/ICEngTechnol.2017.8308186.
\bibitem{b12}S. Thokrairak, K. Thibuy and P. Jitngernmadan, "Valuable Waste Classification Modeling based on SSD-MobileNet," 2020 - 5th International Conference on Information Technology (InCIT), 2020, pp. 228-232, doi: 10.1109/InCIT50588.2020.9310928.
\bibitem{b13}Youpeng Yu, , and Ryan Grammenos. "Towards artificially intelligent recycling Improving image processing for waste classification.", arXiv:2108.06274 [cs.CV],(2021).
\bibitem{b14}K. Ahmad, K. Khan and A. Al-Fuqaha, "Intelligent Fusion of Deep Features for Improved Waste Classification," in IEEE Access, vol. 8, pp. 96495-96504, 2020, doi: 10.1109/ACCESS.2020.2995681.
\bibitem{b15}S. Varudandi, R. Mehta, J. Mahetalia, H. Parmar and K. Samdani, "A Smart Waste Management and Segregation System that Uses Internet of Things, Machine Learning and Android Application," 2021 6th International Conference for Convergence in Technology (I2CT), 2021, pp. 1-6, doi: 10.1109/I2CT51068.2021.9418125.
\bibitem{b16}C. Srinilta and S. Kanharattanachai, "Municipal Solid Waste Segregation with CNN," 2019 5th International Conference on Engineering, Applied Sciences and Technology (ICEAST), 2019, pp. 1-4, doi: 10.1109/ICEAST.2019.8802522.
\bibitem{b17}Kumar, S.; Yadav, D.; Gupta, H.; Verma, O.P.; Ansari, I.A.; Ahn, C.W. A Novel YOLOv3 Algorithm-Based Deep Learning Approach for Waste Segregation: Towards Smart Waste Management. Electronics 2021, 10, 14. https://doi.org/10.3390/electronics10010014
\bibitem{b18}S. R., R. P., V. S., K. R. and G. M., "Deep Learning based Smart Garbage Classifier for Effective Waste Management," 2020 5th International Conference on Communication and Electronics Systems (ICCES), 2020, pp. 1086-1089, doi: 10.1109/ICCES48766.2020.9137938.
\bibitem{b19}Pablo Ruiz, "Understanding and visualizing ResNets", 2018, [Online]. Available: https://towardsdatascience.com/understanding-and-visualizing-resnets-442284831be8
\bibitem{b20}Abhay Parashar, "Vgg 16 Architecture, Implementation and Practical Use", 2020, [Online]. Available: https://medium.com/pythoneers/VGG16-architecture-implementation-and-practical-use-e0fef1d14557
\end{thebibliography}
\end{document}